\documentclass[letterpaper, 10 pt, conference]{ieeeconf}
\usepackage{xurl}
\usepackage{amsmath}
\usepackage{makecell}
\usepackage{multirow}
\usepackage{booktabs}
\usepackage{cite}
\usepackage{ulem}
\usepackage{marvosym}
\usepackage{amssymb}
\usepackage{mathrsfs}
\usepackage{algorithm}
\usepackage{graphicx}
\usepackage{subfig}

\usepackage[colorlinks=true, urlcolor=blue, linkcolor=red]{hyperref}

\usepackage{xcolor}
\usepackage{tikz}

\newcolumntype{L}[1]{>{\raggedright\arraybackslash}p{#1}}
\newcolumntype{C}[1]{>{\centering\arraybackslash}p{#1}}
\newcolumntype{R}[1]{>{\raggedleft\arraybackslash}p{#1}}

\newcommand\normx[1]{\left\Vert#1\right\Vert}

\newcommand{\eg}{\textit{e.g.}}

\newcommand{\etal}{\textit{et al.}}

\hypersetup{
    colorlinks=true,
    linkcolor=blue,
    filecolor=black,      
    urlcolor=purple,
    citecolor=red
}
\definecolor{bariros}{rgb}{0.8000, 0.8000, 1.0000}
\definecolor{barours}{rgb}{1.0000, 0.6000, 0.6000}
\newcommand{\bariros}{\raisebox{1.2pt}{\tikz{\draw[-,bariros,solid,line width = 4pt](0,0) -- (3mm,0);}}}
\newcommand{\barours}{\raisebox{1.2pt}{\tikz{\draw[-,barours,solid,line width = 4pt](0,0) -- (3mm,0);}}}

\IEEEoverridecommandlockouts
\title{\LARGE \bf
Dive Deeper into Rectifying Homography for\\Stereo Camera Online Self-Calibration
}
\author{Hongbo Zhao, Yikang Zhang, Qijun Chen,~\IEEEmembership{Senior Member,~IEEE}, and Rui Fan\textsuperscript{\Letter},~\IEEEmembership{Senior Member,~IEEE}
\thanks{
This research was supported by the National Key R\&D Program of China under Grant 2020AAA0108100, the National Natural Science Foundation of China under Grant 62233013, the Science and Technology Commission of Shanghai Municipal under Grant 22511104500, the Fundamental Research Funds for the Central Universities, and Xiaomi Young Talents Program
(\emph{\Letter\ Corresponding author: Rui Fan}).}
\thanks{All authors are with the College of Electronics \& Information Engineering, Shanghai Research Institute for Intelligent Autonomous Systems, the State Key Laboratory of Intelligent Autonomous Systems, and Frontiers Science Center for Intelligent Autonomous Systems, Tongji University, Shanghai 201804, China (e-mails: \{hongbozhao, yikangzhang, qjchen\}@tongji.edu.cn, {rui.fan@ieee.org}).
}
}

\begin{document}
	
\maketitle
\thispagestyle{empty}
\pagestyle{empty}
	
\begin{abstract}
Accurate estimation of stereo camera extrinsic parameters is crucial to guarantee the performance of stereo matching algorithms. In prior arts, the online self-calibration of stereo cameras has commonly been formulated as a specialized visual odometry problem, without taking into account the principles of stereo rectification. In this paper, we first delve deeply into the concept of rectifying homography, which serves as the cornerstone for the development of our novel stereo camera online self-calibration algorithm, for cases where only a single pair of images is available. Furthermore, we introduce a simple yet effective solution for global optimum extrinsic parameter estimation in the presence of stereo video sequences. Additionally, we emphasize the impracticality of using three Euler angles and three components in the translation vectors for performance quantification. Instead, we introduce four new evaluation metrics to quantify the robustness and accuracy of extrinsic parameter estimation, applicable to both single-pair and multi-pair cases. Extensive experiments conducted across indoor and outdoor environments using various experimental setups validate the effectiveness of our proposed algorithm. The comprehensive evaluation results demonstrate its superior performance in comparison to the baseline algorithm. Our source code, demo video, and supplement are publicly available at \url{mias.group/StereoCalibrator}.
\end{abstract}

\section{Introduction}
\label{sec.intro}

Stereo vision is a fundamental robot perception technique, widely used to acquire dense depth information from a pair of synchronized images \cite{fan2021graph, wang2021pvstereo, fan2021rethinking, wu2024s, fan2018road}. Stereo camera calibration is typically carried out in an offline fashion using a checkerboard pattern, producing the intrinsic matrices $\boldsymbol{K}_l$ and $\boldsymbol{K}_r$ for the left and right cameras, respectively, as well as the extrinsic matrix $\boldsymbol{P}$, defining the relative transformation between the two cameras as follows:
\begin{equation}
    \tilde{\boldsymbol{p}}^C_r=\boldsymbol{P}\tilde{\boldsymbol{p}}^C_l=
    \begin{bmatrix}
        \begin{array}{ll}
        \boldsymbol{R} & \boldsymbol{t}\\
        \boldsymbol{0}^\top & 1
        \end{array}
    \end{bmatrix}
    \tilde{\boldsymbol{p}}^C_l,
    \label{eq.pl2pr}
\end{equation}
where $\boldsymbol{R}\in\text{SO(3)}$ represents the rotation matrix, $\boldsymbol{t}$ denotes the translation vector, $\boldsymbol{0}$ represents a vector of zeros, and $\tilde{\boldsymbol{p}}^C_l$ and $\tilde{\boldsymbol{p}}^C_r$ denote the homogeneous coordinates of $\boldsymbol{p}^C_l=[x_l, y_l, z_l]^{\top}$ and $\boldsymbol{p}^C_r=[x_r, y_r, z_r]^{\top}$ in the left and right camera coordinate systems, respectively. By estimating the rectifying homography based on $\boldsymbol{R}$ and $\boldsymbol{t}$, and pre-processing the raw stereo image pairs, stereo matching is simplified as a 1-D dense correspondence search problem \cite{fan2023computer}. 

Research conducted with various types of stereo cameras has shown that $\boldsymbol{K}_l$ and $\boldsymbol{K}_r$ remain relatively stable even in challenging conditions, \eg, crashes or prolonged mechanical vibrations \cite{ling2016high}. In contrast, $\boldsymbol{R}$ and $\boldsymbol{t}$ can undergo significant changes even with moderate shocks or during extended operations in environments with vibrations, such as those experienced by aerial robots \cite{song2023cnn}. A minor error in the extrinsic calibration of the stereo camera can lead to a non-negligible disparity error \cite{dexheimer2022information}. This could have serious consequences, particularly in safety-critical applications, such as autonomous driving and mobile robot navigation, where miscalculating the distance to a nearby obstacle could lead to a collision \cite{yang2019drivingstereo}. Therefore, stereo camera online self-calibration is essential to ensure the accuracy and reliability of the depth information obtained through stereo matching, making it a cornerstone technology for any robotics system that relies on precise 3-D environmental understanding.

In this paper, we introduce a novel stereo camera online self-calibration algorithm for extrinsic parameter estimation. Our algorithm is developed based on the principles of stereo rectification. Rather than treating stereo camera online self-calibration as a specialized visual odometry problem, where the left and right images are treated as successive video frames and an energy function w.r.t. $\boldsymbol{R}$ and $\boldsymbol{t}$ is minimized, we propose to rotate the left and right camera coordinate systems using two independent rotation matrices $\boldsymbol{R}_l=[\boldsymbol{r}_{l,1}, \boldsymbol{r}_{l,2}, \boldsymbol{r}_{l,3}]^{\top}$ and $\boldsymbol{R}_r=[\boldsymbol{r}_{r,1}, \boldsymbol{r}_{r,2}, \boldsymbol{r}_{r,3}]^{\top}$, respectively, and accumulate the residuals only in the vertical direction. This strategy effectively prevents the emergence of a positive semidefinite coefficient matrix during the optimization process for the translation vector, consequently improving the algorithm's ability to handle initial estimations. 

Furthermore, instead of resorting to the computationally intensive visual odometry back-end optimization algorithms to estimate global optimum extrinsic parameters $\boldsymbol{R}^*$ and $\boldsymbol{t}^*$, we propose a simple yet effective solution by introducing a novel energy function that measures the cosine similarities between a collection of given normalized vectors and the optimum one. Its closed-form solution can be easily derived. This energy function is employed to estimate the global optimum rotation axis and the translation vector. 

Moreover, in \cite{ling2016high}, the mean values and standard deviations of three Euler angles (computed from the rotation matrix) as well as the three components in the translation vectors are used as the evaluation metrics. However, these evaluation metrics are impractical due to the interdependency among the individual components with the other two. Therefore, we propose four new evaluation metrics to comprehensively quantify both the robustness and accuracy of extrinsic parameter estimation for both single-pair and multi-pair cases.

Through extensive experiments conducted on a large-scale dataset containing over 6K pairs of real-world stereo images collected from both indoor and outdoor environments with respect to different extrinsic parameters, as well as on two public datasets, KITTI 2015 \cite{menze2015joint} and Middlebury 2021 \cite{scharstein2014high}, perturbed in different directions, we demonstrate the effectiveness of our proposed algorithm and its superior performance compared to the baseline algorithm. 

\section{Related Works}
\label{sec.related_work}

Stereo camera online self-calibration is an important yet under-investigated problem in the current body of literature. This section reviews existing explicit programming-based approaches. Most prior arts treat this task as a specialized form of visual odometry, in which $\boldsymbol{R}$ and $\boldsymbol{t}$ for each stereo pair are obtained via the decomposition of the essential matrix $\boldsymbol{E}=[\boldsymbol{t}]_{\times}\boldsymbol{R}$. Subsequently, $\boldsymbol{R}^*$ and $\boldsymbol{t}^*$ are derived using back-end optimization techniques, \eg, bundle adjustment (BA) \cite{triggs2000bundle} and Kalman filter (KF) \cite{chen2023kalman}. For example, Mueller and Wuensche \cite{mueller2016continuous} presented an algorithm to estimate the transformation from the vehicle coordinate system to the left and right camera coordinate systems, respectively, as well as the relative transformation between two cameras. Salient correspondences observed by both cameras are continuously tracked within a 3-D space. An unscented Kalman filter (UKF) algorithm is subsequently applied to recursively refine the extrinsic parameters of the stereo camera and update the 3-D coordinates of all observed points. In their follow-up study \cite{mueller2017continuous}, Mueller and Wuensche introduced a sequential, block-wise filtering approach,  where measurements corresponding to the same 3-D point observed by both cameras are jointly filtered. Additionally, they leveraged an extended Kalman filter (EKF) to yield $\boldsymbol{R}^*$ and $\boldsymbol{t}^*$. 

Rather than directly estimating $\boldsymbol{R}$ and $\boldsymbol{t}$, Dang {\etal} \cite{dang2009continuous} proposed to separately rotate the left and right camera coordinate systems using two independent rotation matrices $\boldsymbol{R}_l$ and $\boldsymbol{R}_r$. Additionally, they explored the impact of calibration errors on 3-D reconstruction and designed a continuous, recursive refinement technique based on iterative EKF to simultaneously update both intrinsic and extrinsic parameters of the stereo camera. Subsequently, Hansen {\etal} \cite{hansen2012online} extended the work \cite{dang2009continuous} by formulating and solving the following optimization problem based on stereo rectification for $\boldsymbol{R}_l$ and $\boldsymbol{R}_r$ estimation:
\begin{equation}
    \underset{\boldsymbol{R}_l,\boldsymbol{R}_r}{\arg\min} \sum_i \normx{\frac{\boldsymbol{r}_{l,2}^{\top}\boldsymbol{p}^{C}_{l,i}}{\boldsymbol{r}_{l,3}^{\top}\boldsymbol{p}^{C}_{l,i}}-\frac{\boldsymbol{r}_{r,2}^{\top}\boldsymbol{p}^{C}_{r,i}}{\boldsymbol{r}_{r,3}^{\top}\boldsymbol{p}^{C}_{r,i}}}^2_2,
\end{equation}
where $\boldsymbol{p}^{C}_{l,i}$ and $\boldsymbol{p}^{C}_{r,i}$ denote the $i$-th pair of corresponding 3-D points in the left and right camera coordinate systems, respectively. $\boldsymbol{R}^*$ and $\boldsymbol{t}^*$ are also obtained via KF. 

The aforementioned studies merely focus on estimating $\boldsymbol{R}$ and $\boldsymbol{t}$ in single-pair cases and mainly seek global optimum solutions $\boldsymbol{R}^*$ and $\boldsymbol{t}^*$ using Kalman filters. Ling and Shen \cite{ling2016high} proposed to solve the following optimization problem based on the epipolar constraint:
\begin{equation}
    \underset{\boldsymbol{R},\boldsymbol{t}}{\arg\min} \sum_i \normx{{\boldsymbol{p}^C_{l,i}}^\top [\boldsymbol{t}]_\times\boldsymbol{R} \boldsymbol{p}^C_{r,i}}^2_2.
\end{equation} 
However, their study does not extend to the estimation of $\boldsymbol{R}^*$ and $\boldsymbol{t}^*$. Inspired by prior works \cite{dang2009continuous, hansen2012online, ling2016high}, we introduce a novel stereo camera online self-calibration algorithm for both single-pair and multi-pair cases. 

\section{Methodology}
\label{sec.methodology}

\subsection{Dive deeper into rectifying homography}
\label{sec.preliminaries}
In this subsection, we first provide readers with mathematical preliminaries of stereo rectification, which is typically undertaken prior to performing stereo matching. A pair of correspondences $\boldsymbol{p}_l = [u_l, v_l]^{\top}$ and $\boldsymbol{p}_r = [u_r, v_r]^{\top}$, respectively in the left and right stereo images, can be associated with their camera coordinates $\boldsymbol{p}^C_l$ and $\boldsymbol{p}^C_r$ using their known intrinsic matrices $\boldsymbol{K}_l$ and $\boldsymbol{K}_r$ as follows:
\begin{equation}
	z_l \tilde{\boldsymbol{p}}_l=\boldsymbol{K}_l\boldsymbol{p}^C_l, \ \ z_r \tilde{\boldsymbol{p}}_r=\boldsymbol{K}_r\boldsymbol{p}^C_r,
 \label{eq.cam_l}
\end{equation}
where $\tilde{\boldsymbol{p}}_{l,r}$ represents the homogeneous coordinates of $\boldsymbol{p}_{l,r}$. The stereo rectification process that reprojects left and right image planes onto a common plane parallel to the stereo rig baseline can be expressed as follows:
\begin{equation}
    \begin{cases}
        \begin{aligned}
            z'_l \tilde{\boldsymbol{p}}'_l &=\boldsymbol{K}{\boldsymbol{p}^{C}_l}{'}
            =\boldsymbol{K}\boldsymbol{R}_l \boldsymbol{p}^{C}_l
            =z_l\boldsymbol{K}\boldsymbol{R}_l\boldsymbol{K}^{-1}_l\tilde{\boldsymbol{p}}_l,
        \end{aligned}
        \\
        \begin{aligned}
            z'_r \tilde{\boldsymbol{p}}'_r &=\boldsymbol{K}{\boldsymbol{p}^{C}_r}{'}
            =\boldsymbol{K}\boldsymbol{R}_r \boldsymbol{p}^{C}_r
            =z_r\boldsymbol{K}\boldsymbol{R}_r \boldsymbol{K}^{-1}_r\tilde{\boldsymbol{p}}_r,
        \end{aligned}
    \end{cases}
\label{eq.rectification_lr}
\end{equation}
where ${\boldsymbol{p}^{C}_l}{'}=[x'_l,y'_l,z'_l]^{\top}$ and ${\boldsymbol{p}^{C}_r}{'}=[x'_r,y'_r,z'_r]^{\top}$ represent the transformed coordinates of $\boldsymbol{p}^C_l$ and $\boldsymbol{p}^C_r$ in the new left and right camera coordinate systems, respectively, $\boldsymbol{p}'_l = [u'_l, v'_l]^{\top}$ and $\boldsymbol{p}'_r = [u'_r, v'_r]^{\top}$ denote the projections of ${\boldsymbol{p}^{C}_l}{'}$ and ${\boldsymbol{p}^{C}_r}{'}$ in the rectified left and right images, respectively, 
and $\boldsymbol{K}$ is the newly defined camera intrinsic matrix. Due to the inherent scale ambiguity in the uncalibrated stereo geometry \cite{hartley2003multiple}, we treat $\boldsymbol{t}$ in (\ref{eq.pl2pr}) as a unit vector and obtain the following expression\footnote{An identity matrix $\boldsymbol{I} =[\boldsymbol{i}_1, \boldsymbol{i}_2, \boldsymbol{i}_3]$.}:
\begin{equation}
    {\boldsymbol{p}^{C}_l}{'} = {\boldsymbol{p}^{C}_r}{'} + \boldsymbol{i}_1.
 \label{eq.Rt'}
\end{equation}
Combining (\ref{eq.rectification_lr}), (\ref{eq.Rt'}) and (\ref{eq.pl2pr}) results in:

\begin{equation}
\begin{cases}
        \begin{aligned}
            \boldsymbol{R} = \boldsymbol{R}_r^{-1}\boldsymbol{R}_l,
        \end{aligned}
        \\
        \begin{aligned}
            \boldsymbol{t} = -\boldsymbol{r}_{r,1}.
        \end{aligned}
\end{cases}
\label{eq.Rt}
\end{equation}
Drawing upon the principles of stereo rectification \cite{trucco1998introductory}, we yield the expression of $\boldsymbol{R}_r$ as follows:
\begin{equation}
\boldsymbol{R}_r=
\big[
-\boldsymbol{t}, \ \  \boldsymbol{i}_3 \times \boldsymbol{r}_{r,1}, \ \ \boldsymbol{r}_{r,1} \times \boldsymbol{r}_{r,2}
\big]^\top.
    \label{eq.R_R}
\end{equation}
Therefore, the conventional stereo camera calibration problem, which involves the estimation of $\boldsymbol{R}$ and $\boldsymbol{t}$, can be redefined as a novel problem focused on estimating $\boldsymbol{R}_l$ and $\boldsymbol{R}_r$. As a pair of well-rectified stereo images adhere to the condition $v'_l=v'_r$, we can deduce the following constraint:
\begin{equation}
    \frac{\boldsymbol{r}_{l,2}^{\top} \boldsymbol{p}^{C}_l}{\boldsymbol{r}_{l,3}^{\top}\boldsymbol{p}^{C}_l}=\frac{\boldsymbol{r}_{r,2}^{\top}\boldsymbol{p}^{C}_r}{\boldsymbol{r}_{r,3}^{\top}\boldsymbol{p}^{C}_r},
    \label{eq.error}
\end{equation}
which is used in our method to formulate the energy function. 
\subsection{Energy function and its solution for single-pair cases}
\label{sec.single_pair}
Based on the stereo rectification constraint introduced in (\ref{eq.R_R}) and (\ref{eq.error}), we define an $(N+1)$-entry vector
\begin{equation}
    \boldsymbol{e}(\boldsymbol{\theta}_l, \boldsymbol{\theta}_r) = [e_0,e_1,\dots,e_{N}]^\top,
\end{equation}
where 
\begin{equation}
    \begin{cases}
        \begin{aligned}
            e_{0}(\boldsymbol{\theta}_l, \boldsymbol{\theta}_r) = \boldsymbol{i}_3^{\top}\boldsymbol{r}_{r,2} = \boldsymbol{i}_{2}^{\top} \exp([\boldsymbol{\theta}_r]_{\times})\boldsymbol{i}_{3},
        \end{aligned}
        \\
        \begin{aligned}
            e_{i}(\boldsymbol{\theta}_l, \boldsymbol{\theta}_r) &= \frac{\boldsymbol{r}_{l,2}^{\top}\boldsymbol{p}^{C}_{l,i}}{\boldsymbol{r}_{l,3}^{\top}\boldsymbol{p}^{C}_{l,i}}-\frac{\boldsymbol{r}_{r,2}^{\top}\boldsymbol{p}^{C}_{r,i}}{\boldsymbol{r}_{r,3}^{\top}\boldsymbol{p}^{C}_{r,i}} = \frac{\boldsymbol{i}_{2}^{\top} \exp([\boldsymbol{\theta}_l]_{\times})\boldsymbol{p}^{C}_{l,i}}{\boldsymbol{i}_{3}^{\top} \exp([\boldsymbol{\theta}_l]_{\times})\boldsymbol{p}^{C}_{l,i}}
            \\ & -\frac{\boldsymbol{i}_{2}^{\top} \exp([\boldsymbol{\theta}_r]_{\times})\boldsymbol{p}^{C}_{r,i}}{\boldsymbol{i}_{3}^{\top} \exp([\boldsymbol{\theta}_r]_{\times})\boldsymbol{p}^{C}_{r,i}}, \ \ \ i\in[1,N]\cap\mathbb{Z},
        \end{aligned}
    \end{cases}
\end{equation}
and $\boldsymbol{\theta}_l$ and $\boldsymbol{\theta}_r$ denote the rotation vectors corresponding to $\boldsymbol{R}_l$ and $\boldsymbol{R}_r$, respectively. $e_1$ to $e_{N}$ represent the residuals between $v'_l$ and $v'_r$. Therefore, we can formulate the energy function for stereo camera self-calibration as follows:
\begin{equation}
\begin{split}
E&=\normx{\boldsymbol{e}(\boldsymbol{\theta}_l, \boldsymbol{\theta}_r)}^2_2  = \normx{\boldsymbol{i}_{2}^{\top} \exp([\boldsymbol{\theta}_l]_{\times})\boldsymbol{i}_{3}}^2_2 \\ &+
         \sum^{N}_{i=1} \normx{\frac{\boldsymbol{i}_{2}^{\top} \exp([\boldsymbol{\theta}_l]_{\times})\boldsymbol{p}^{C}_{l,i}}{\boldsymbol{i}_{3}^{\top} \exp([\boldsymbol{\theta}_l]_{\times})\boldsymbol{p}^{C}_{l,i}}-\frac{\boldsymbol{i}_{2}^{\top} \exp([\boldsymbol{\theta}_r]_{\times})\boldsymbol{p}^{C}_{r,i}}{\boldsymbol{i}_{3}^{\top} \exp([\boldsymbol{\theta}_r]_{\times})\boldsymbol{p}^{C}_{r,i}}}^2_2.
\end{split}
\label{eq.E}
\end{equation}
The optimum $\boldsymbol{R}_l$ and $\boldsymbol{R}_r$ can be yielded by minimizing (\ref{eq.E}). To this end, we formulate the first-order Taylor expansion of ${e}_i(\boldsymbol{\theta}_l, \boldsymbol{\theta}_r)$ as follows:
\begin{equation}
    e_i(\hat{\boldsymbol{\theta}}_l+\delta \boldsymbol{\theta}_l, \hat{\boldsymbol{\theta}}_r+\delta \boldsymbol{\theta}_r) \approx e_i(\hat{\boldsymbol{\theta}}_l , \hat{\boldsymbol{\theta}}_r) + \boldsymbol{J}_{i}(\hat{\boldsymbol{\theta}}_l , \hat{\boldsymbol{\theta}}_r)^\top \delta\boldsymbol{\theta}, 
    \label{eq.taylor}
\end{equation}
where $\hat{\boldsymbol{\theta}}= [\hat{\boldsymbol{\theta}}_l^\top, \hat{\boldsymbol{\theta}}_r^\top]^\top$ is the current estimate, $\delta\boldsymbol{\theta} = [\delta \boldsymbol{\theta}^{\top}_l, \delta \boldsymbol{\theta}^{\top}_r ]^\top $ is the increment, 
\begin{equation}
    \begin{cases}
        \begin{aligned}
            \boldsymbol{J}_{0}(\boldsymbol{\theta}_l , \boldsymbol{\theta}_r) = [\boldsymbol{0}^\top, \frac{\partial \boldsymbol{r}_{r,23}}{\partial \boldsymbol{\theta}_r^\top}]^\top = [0,0,0,\boldsymbol{i}_2^{\top}[\boldsymbol{R}_r\boldsymbol{i}_3]_{\times}]^\top,
        \end{aligned}
\\
        \begin{aligned}
        \boldsymbol{J}_{i}(\boldsymbol{\theta}_l , \boldsymbol{\theta}_r) = [ {\frac{\partial e_i}{\partial \boldsymbol{\theta}_l^\top}} , {\frac{\partial e_i}{\partial \boldsymbol{\theta}_r^\top}} ]^\top,\ \ i\in[1,N]\cap\mathbb{Z},
        \end{aligned}
    \end{cases}
\end{equation}
and
\begin{equation}
\begin{cases}
        \begin{aligned}
            \frac{\partial e_i}{\partial \boldsymbol{\theta}_l^\top} = \frac{-\boldsymbol{r}_{l,3}^{\top}\boldsymbol{p}^{C}_{l,i}\boldsymbol{i}_2^\top[\boldsymbol{R}_l\boldsymbol{p}^{C}_{l,i}]_{\times} + \boldsymbol{r}_{l,2}^{\top}\boldsymbol{p}^{C}_{l,i}\boldsymbol{i}_3^\top[\boldsymbol{R}_l\boldsymbol{p}^{C}_{l,i}]_{\times}}{(\boldsymbol{r}_{l,3}^{\top}\boldsymbol{p}^{C}_{l,i})^2},
        \end{aligned}
       \\ \begin{aligned}
           \frac{\partial e_i}{\partial \boldsymbol{\theta}_r^\top} = \frac{ \boldsymbol{r}_{r,3}^{\top}\boldsymbol{p}^{C}_{r,i}\boldsymbol{i}_2^\top[\boldsymbol{R}_r\boldsymbol{p}^{C}_{r,i}]_{\times} - \boldsymbol{r}_{r,2}^{\top}\boldsymbol{p}^{C}_{r,i}\boldsymbol{i}_3^\top[\boldsymbol{R}_r\boldsymbol{p}^{C}_{r,i}]_{\times}}{(\boldsymbol{r}_{r,3}^{\top}\boldsymbol{p}^{C}_{r,i})^2}.
        \end{aligned}
    \end{cases} 
\end{equation}
$E^{'}$, the approximate $E$, can be yielded as follows:
\begin{equation}
   E^{'}=\min_{\delta\boldsymbol{\theta}} \sum^{N}_{i=0} \normx{e_i(\hat{\boldsymbol{\theta}}_l , \hat{\boldsymbol{\theta}}_r) + \boldsymbol{J}_i(\hat{\boldsymbol{\theta}}_l , \hat{\boldsymbol{\theta}}_r)^\top \delta\boldsymbol{\theta}}^2_2.
\end{equation}
$\delta\boldsymbol{\theta}$ can be solved using the Levenberg-Marquardt (LM) algorithm \cite{marquardt1963algorithm} as follows:
\begin{equation}
    (\sum^{N}_{i=0} w_{i} {\boldsymbol{J}_i}{\boldsymbol{J}_i}^{\top} + \lambda \boldsymbol{I}) \delta\boldsymbol{\theta} = - \sum^{N}_{i=0} w_{i} {\boldsymbol{J}_i}\boldsymbol{e}_i,
\end{equation}
where $\lambda$ is the damping factor used in the LM algorithm, \begin{align}
w_i = 
\begin{cases}
         1,    & |{e}_i| \le c_t \\
         {c_t}/{|{e}_i|},     & {|{e}_i|} > c_t
\end{cases}
\label{eq.fu_fv_wrt_uv}
\end{align}
is the weight assigned to the $i$-th pair of correspondences \cite{ling2016high}, and $c_t$ represents a pre-defined Huber norm \cite{huber1992robust} threshold.
We can then update $\hat{\boldsymbol{\theta}}_l$ and $\hat{\boldsymbol{\theta}}_l$ through:
\begin{equation}
    \begin{cases}
        \begin{aligned}
        {\hat{\boldsymbol{R}}}_l \gets \exp([\delta {\boldsymbol{\theta}_l}]_{\times}) {\hat{\boldsymbol{R}}}_l, \ \ 
        \hat{\boldsymbol{\theta}}_l \gets \log(\hat{\boldsymbol{R}}_l),
        \end{aligned}
        \\
         \begin{aligned}
       \hat{\boldsymbol{R}}_r \gets \exp([\delta {\boldsymbol{\theta}_r}]_{\times})\hat{\boldsymbol{R}}_r, \ \hat{\boldsymbol{\theta}}_r \gets \log(\hat{\boldsymbol{R}}_r).
        \end{aligned}
    \end{cases}   
\end{equation}
When $\normx{\delta\boldsymbol{\theta}}_2$ is considered to be insufficiently small, the update process for $\hat{\boldsymbol{R}}_l$, $\hat{\boldsymbol{R}}_r$, $\hat{\boldsymbol{\theta}}_l$, and $\hat{\boldsymbol{\theta}}_r$ terminates. The extrinsic parameters $\boldsymbol{R}$ and $\boldsymbol{t}$ of the stereo camera can, therefore, be obtained by substituting $\hat{\boldsymbol{R}}_l$ and $\hat{\boldsymbol{R}}_r$ into (\ref{eq.Rt}).

\subsection{Global optimization for multi-pair cases}
\label{sec.Global optimization for multi-pair cases}
Upon acquiring the extrinsic parameters $\boldsymbol{R}^{k}$ and $\boldsymbol{t}^{k}$ for each pair of stereo images, with the superscript $k$ denoting the $k$-th stereo image pair, it becomes feasible to deduce the global optimum extrinsic parameters $\boldsymbol{R}^{*}$ and $\boldsymbol{t}^{*}$ for the stereo camera. Given that $\boldsymbol{t}^{k}$ are normalized, they can be projected onto a sphere within a confined region. Similar distributions are also witnessed for the rotation axes $\boldsymbol{v}^{k}$, which can be computed using the Rodrigues' rotation formula \cite{rodrigues1840lois} as follows:
\begin{equation}
    \begin{cases}
        \begin{aligned}
        s^k=\arccos(\frac{\text{tr}(\boldsymbol{R}^{k})-1}{2}),
        \end{aligned}
        \\
        \begin{aligned}
        [\boldsymbol{v}^{k}]_{\times} = \frac{\boldsymbol{R}^{k}-{\boldsymbol{R}^{k}}^\top}{2\sin s^k},
        \end{aligned}
    \end{cases}   
\end{equation}
where $s^k$ is the rotation angle. The optimum translation vector $\boldsymbol{t}^{*}$ and rotation axis $\boldsymbol{v}^{*}$ can, therefore, be determined by finding the position on the sphere, where the projections distribute most intensively \cite{fan2019pothole}. This can be achieved through the following global optimization process using $M$ groups of extrinsic parameters $\boldsymbol{R}^{k}$ and $\boldsymbol{t}^{k}$:
\begin{equation}
    \begin{cases}
        \begin{aligned}
        \boldsymbol{t}^* = \underset{{\boldsymbol{t}^*}}{{\arg\max}}\sum^{M}_{k=1} {\boldsymbol{t}^{k}}^\top\boldsymbol{t}^{*}= \frac{\sum^{M}_{k=1}\boldsymbol{t}^{k}}{\normx{\sum^{M}_{k=1}\boldsymbol{t}^{k}}_2},
        \end{aligned}
        \\
        \begin{aligned}
        \boldsymbol{v}^* = \underset{{\boldsymbol{v}^*}}{{\arg\max}}\sum^{M}_{k=1} {\boldsymbol{v}^{k}}^\top\boldsymbol{v}^{*}= \frac{\sum^{M}_{k=1}\boldsymbol{v}^{k}}{\normx{\sum^{M}_{k=1}\boldsymbol{v}^{k}}_2}.
        \end{aligned}
    \end{cases}   
\end{equation}
The optimum rotation matrix $\boldsymbol{R}^{*}$ can, therefore, be obtained as follows:
\begin{equation}
    \boldsymbol{R}^{*} = \exp([\boldsymbol{\theta}^{*}]_\times)= \exp([s^{*}\boldsymbol{v}^{*}]_\times),
\end{equation}
where $s^{*}$ is determined using the central tendency measurement of all rotation angles $s^{k}$, and $\boldsymbol{\theta}^{*}$ represents the global optimum rotation vector.

\section{Experimental Results}
\label{sec.experiments}

\subsection{Experiment setup and implementation details}
\label{sec.exp_setup}

\begin{figure}[t!]
\vspace{+0.5em}
\centering
\includegraphics[width=0.46\textwidth]{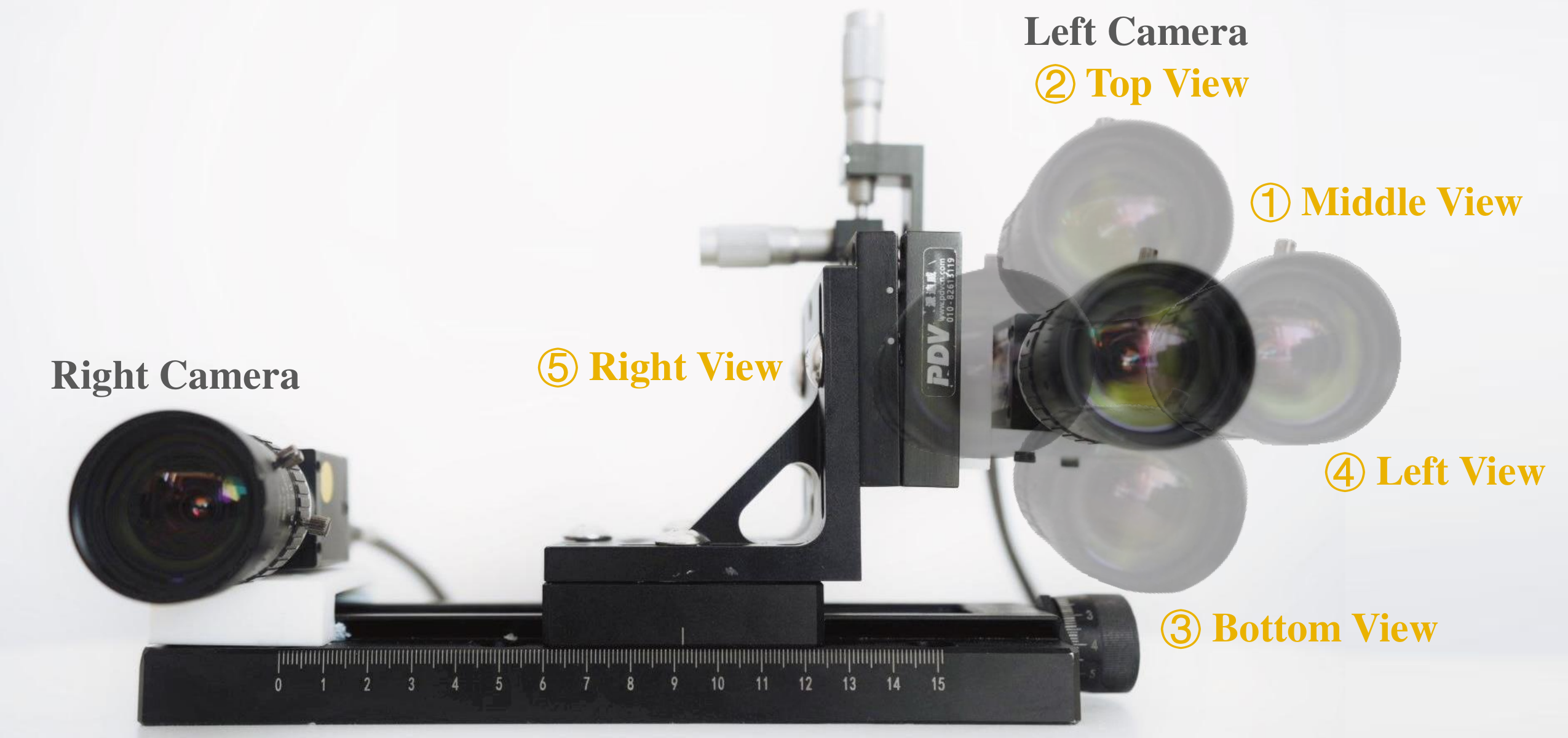}
\caption{The experimental setup with the left camera mounted at five different viewpoints.}
\label{fig.exp_setup}
\vspace{-1.8em}
\end{figure}

In our experiments, we used two MV-SUA202GC global-shutter CMOS cameras from MindVision to collect data (around 6K stereo pairs with a resolution of 1,920$\times$1,200 pixels) from both indoor and outdoor environments. We achieved camera hardware synchronization by utilizing a 20Hz sync signal supplied by an FPGA, in conjunction with an external power source delivering 24V. Rather than placing the two cameras at relatively ideal positions similar to those in \cite{ling2016high}, we mounted the left camera at five different viewpoints (middle, top, bottom, left, and right views) to conduct a comprehensive evaluation of our algorithm's performance, as illustrated in Fig. \ref{fig.exp_setup}. The intrinsic parameters of each camera are acquired through offline calibration with a checkerboard using the algorithm proposed in \cite{zhang2000flexible} and are assumed to remain consistent throughout the experiments.

Furthermore, we used two public stereo matching datasets, KITTI 2015 \cite{menze2015joint} and Middlebury 2021 \cite{scharstein2014high}, to further quantify the performance of our algorithm. Similarly, we manually created four additional viewpoints (top, bottom, left, and right views) with a rotation angle of $5^\circ$. 

\subsection{Evaluation metrics}
\label{sec.evaluation_metrics}

Ling and Shen \cite{ling2016high} utilized the mean values and standard deviations of three Euler angles (computed from the rotation matrix) and the three components in the translation vectors as the evaluation metrics for performance quantification. Nevertheless, these evaluation metrics are impractical due to the interdependency among the individual components with the other two. Therefore, we propose the following four new evaluation metrics to comprehensively quantify both the robustness and accuracy of extrinsic parameter estimation for both single-pair and multi-pair cases.
\begin{itemize}
\item $e_{\boldsymbol{t}}$ representing the angular error between the estimated and ground-truth global optimum translation vectors:
\begin{equation}
    e_{\boldsymbol{t}} = \arccos(\boldsymbol{t}^\top {\boldsymbol{t}^*}),
    \label{eq.e_t}
\end{equation}
\item $e_{\boldsymbol{\theta}}$ representing the distance between the estimated and ground-truth global optimum rotation vectors:
\begin{equation}
    e_{\boldsymbol{\theta}} = \normx{ \boldsymbol{\theta}^{*}-\boldsymbol{\theta} }_2,
    \label{eq.e_theta}
\end{equation}
\item $\sigma_{\boldsymbol{t}}$ representing the standard deviation of angular errors between the estimated (using a single stereo pair) and the ground-truth global optimum translation vectors:
\begin{equation}
\sigma_{\boldsymbol{t}} = 
\sqrt{
\frac{1}{M}
\sum_{k=1}^{M} 
\bigg(
\arccos(\boldsymbol{t}^\top {\boldsymbol{t}^{k}} )
\bigg)^2,
\label{eq.sigma_t}
}
\end{equation}
\item $\sigma_{\boldsymbol{\theta}}$ representing the standard deviation of distances between the estimated (using a single stereo pair) and the ground-truth global optimum rotation vectors:
\begin{equation}
\sigma_{\boldsymbol{\theta}}= \sqrt{\frac{1}{M} \sum^{M}_{k=1}{\normx{\boldsymbol{\theta}^{k}-\boldsymbol{\theta}}}_2^2},
\label{eq.sigma_theta}
\end{equation}
\end{itemize}
where $\boldsymbol{t}$ and $\boldsymbol{\theta}$ are obtained via stereo camera offline extrinsic calibration with a checkerboard using the algorithm introduced in \cite{zhang2000flexible}. $e_{\boldsymbol{t}}$ and $e_{\boldsymbol{\theta}}$ are used to quantify the accuracy of the global optimum extrinsic parameters obtained using all stereo image pairs, while $ \sigma_{\boldsymbol{t}}$ and $\sigma_{\boldsymbol{\theta}}$ are used to quantify the robustness of single-pair online self-calibration. 

\begin{table*}[t!]
\vspace{+0.5em}
        \fontsize{7.3}{9.6}\selectfont
	\centering
	\caption{
        A quantitative comparison of the global optimum extrinsic parameter estimation between our proposed algorithm and \cite{ling2016high} on our created large-scale dataset.
	}
	\begin{tabular}
	    {c|c|C{1cm} C{1cm}|C{1cm} C{1cm}|C{1cm} C{1cm}|C{1cm} C{1cm}|C{1cm} C{1cm}}
		\toprule
           \multirow{2}{*}{Senario} & \multirow{2}{*}{Algorithm} & \multicolumn{2}{c}{Middle View} & \multicolumn{2}{c}{Top View} & \multicolumn{2}{c}{Bottom View} & \multicolumn{2}{c}{Left View} & \multicolumn{2}{c}{Right View}\\
           \cline{3-12}
            &  & $e_{\boldsymbol{t}}$ (rad) & $e_{\boldsymbol{\theta}} $ (rad) & $e_{\boldsymbol{t}}$ (rad) & $e_{\boldsymbol{\theta}}$ (rad) & $e_{\boldsymbol{t}}$ (rad) & $e_{\boldsymbol{\theta}}$ (rad) & $ e_{\boldsymbol{t}}$ (rad)& $ e_{\boldsymbol{\theta}}$ (rad) & $ e_{\boldsymbol{t}}$ (rad) & $ e_{\boldsymbol{\theta}}$ (rad) \\
		\hline   \hline
		\multicolumn{1}{c|}{\multirow{2}{*}{Indoor 1}}
            & \cite{ling2016high}  &0.2897 & 0.0398 &  0.0274& 0.0102 &  0.1088& 0.0154 & 0.1206 & 0.0173 &  0.0468 & 0.0104   \\
	    &  \textbf{Ours} & \textbf{0.0562} & \textbf{0.0071} & \textbf{0.0053} & \textbf{0.0064} &  \textbf{0.0571} & \textbf{0.0067} &  \textbf{0.0198}& \textbf{0.0150}&  \textbf{0.0061}      & \textbf{0.0026} \\
	\hline
        \multicolumn{1}{c|}{\multirow{2}{*}{Indoor 2}} 
                                                 & \cite{ling2016high} &0.2181 & 0.0165  &0.1297 & 0.0121  &0.1192 & 0.0139  &0.1236 & 0.0163& 0.1029& 0.0194   \\
	                                              &  \textbf{Ours} & \textbf{0.0567}& \textbf{0.0048}  &\textbf{0.0514} & \textbf{0.0039} & \textbf{0.0498} & \textbf{0.0091} & \textbf{0.0338} & \textbf{0.0161} & \textbf{0.0173} & \textbf{0.0121}  \\
    \hline
        \multicolumn{1}{c|}{\multirow{2}{*}{Outdoor 1}}
                                                 & \cite{ling2016high}   & 0.1134 & 0.0225 &0.1578&0.0366  & 0.1394&0.0184  & 0.0318& 0.0074 & 0.1828& 0.0112 \\
	                                              &  \textbf{Ours} & \textbf{0.0238} & \textbf{0.0014}  & \textbf{0.0103} & \textbf{0.0031} & \textbf{0.0106} & \textbf{0.0053} & \textbf{0.0052} & \textbf{0.0038} & \textbf{0.0289} & \textbf{0.0029} \\
    \hline
        \multicolumn{1}{c|}{\multirow{2}{*}{Outdoor 2}} 
                                                 & \cite{ling2016high}  &0.1448 & 0.0087 & 0.2730& 0.0114 & 0.1257& 0.0247 & 0.1349 & 0.0142  & 0.2466 & 0.0521 \\
	                                              &  \textbf{Ours} & \textbf{0.0808} & \textbf{0.0016} & \textbf{0.0337} & \textbf{0.0037} & \textbf{0.0367} & \textbf{0.0023} & \textbf{0.0237} & \textbf{0.0074} & \textbf{0.1308} & \textbf{0.0198}  \\
    \bottomrule
		\end{tabular}
  \label{tab.our_dataset_et_etheta}
  \vspace{-1em}
\end{table*}

\begin{figure*}[t]
\centering
\includegraphics[width=0.99\textwidth]{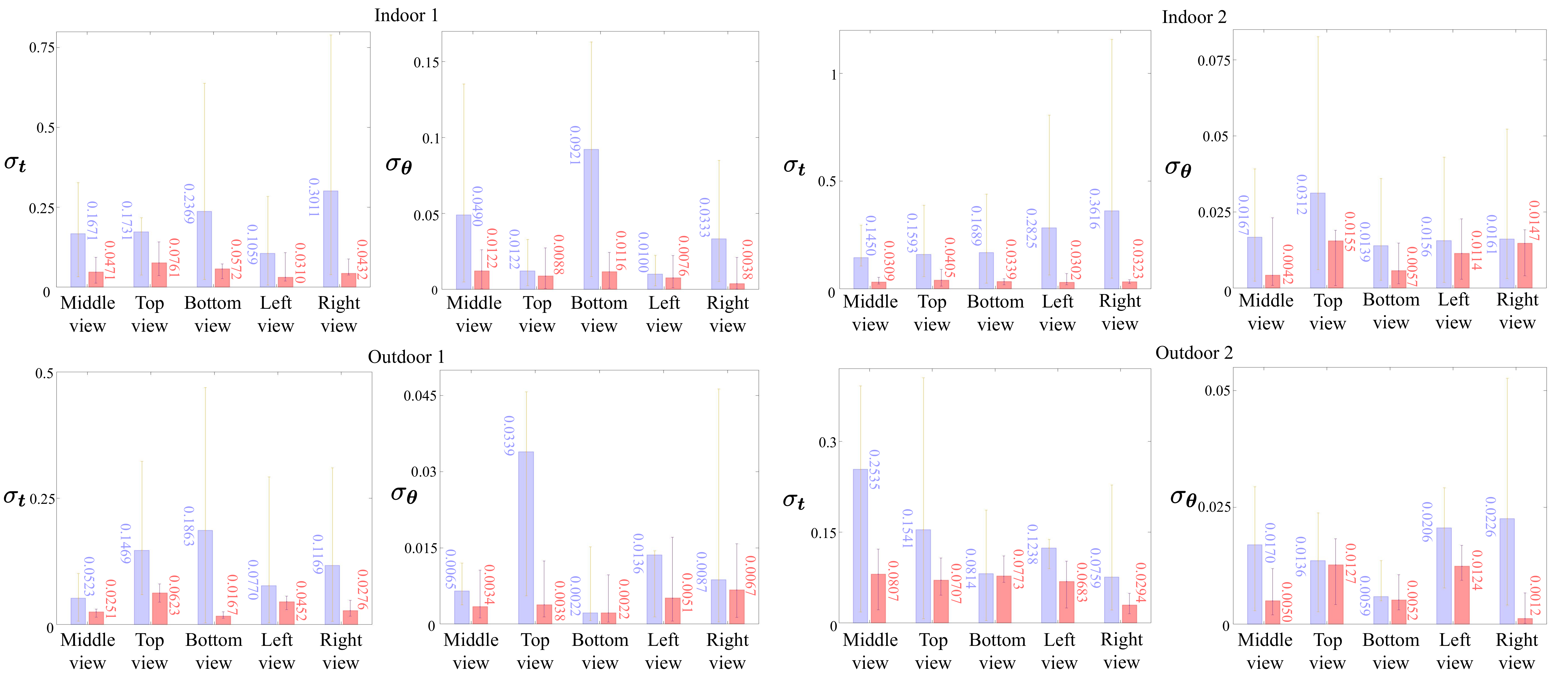}
\caption{
Comparison between \cite{ling2016high} and our proposed algorithm on our created large-scale datasets. {\protect\bariros} presents the results achieved by Ling and Shen \cite{ling2016high}, and {\protect\barours} presents the results achieved by our proposed method. 
}
\label{fig.our_dataset_sigma}
\vspace{-0.8em}
\end{figure*}

\begin{table*}[t!]
        \fontsize{7.3}{9.6}\selectfont
	\centering
	\caption{
        A quantitative comparison of the global optimum extrinsic parameter estimation between our proposed algorithm and \cite{ling2016high} on the KITTI 2015 \cite{menze2015joint} and Middlebury 2021 \cite{scharstein2014high} datasets.
	}
\label{tab.kitti_middleburry_error}
	\begin{tabular}
	    {c|c|C{1cm} C{1cm} | C{1cm} C{1cm} | C{1cm} C{1cm} | C{1cm} C{1cm} | C{1cm} C{1cm}}
		\toprule
           \multirow{2}{*}{Dataset} & \multirow{2}{*}{Algorithm} & \multicolumn{2}{c}{Middle View} & \multicolumn{2}{c}{Top View} & \multicolumn{2}{c}{Bottom View} & \multicolumn{2}{c}{Left View} & \multicolumn{2}{c}{Right View}\\
           \cline{3-12}
            &  & $e_{\boldsymbol{t}}$ (rad) & $e_{\boldsymbol{\theta}}$ (rad) & $e_{\boldsymbol{t}}$ (rad) & $e_{\boldsymbol{\theta}}$ (rad) & $e_{\boldsymbol{t}}$ (rad) & $e_{\boldsymbol{\theta}}$ (rad) & $e_{\boldsymbol{t}}$ (rad) & $e_{\boldsymbol{\theta}}$ (rad) & $e_{\boldsymbol{t}}$ (rad) & $e_{\boldsymbol{\theta}}$ (rad)\\
		\hline   \hline
		\multicolumn{1}{c|}{\multirow{2}{*}{KITTI}}
                                                 & \cite{ling2016high}  & 0.0715 & 0.0030 & 0.1155 & 0.0027 & 0.0600 & 0.0037 & 0.1494 & 0.0053 & 0.0461 & 0.0036  \\
	                                              &  \textbf{Ours} & \textbf{0.0130} & \textbf{0.0014} & \textbf{0.0163} & \textbf{0.0024} & \textbf{0.0322} & \textbf{0.0027} & \textbf{0.0282} & \textbf{0.0027} & \textbf{0.0370} & \textbf{0.0018} \\
	\hline
        \multicolumn{1}{c|}{\multirow{2}{*}{Middleburry}} 
                                                 & \cite{ling2016high}  & 0.0171 & 0.0007 & 0.0185 & 0.0006 & 0.0490 & 0.0086 & 0.0184 & 0.0010 & 0.0200 & \textbf{0.0006}  \\
	                                              &  \textbf{Ours} &  \textbf{0.0084} & \textbf{0.0005} & \textbf{0.0050} & \textbf{0.0004} & \textbf{0.0091} & \textbf{0.0009} & \textbf{0.0094} & \textbf{0.0004} & \textbf{0.0048} & 0.0008 \\
    \bottomrule
		\end{tabular}
  \vspace{-0.5em}
\end{table*}

\begin{figure*}[t]
\centering
\includegraphics[width=0.99\textwidth]{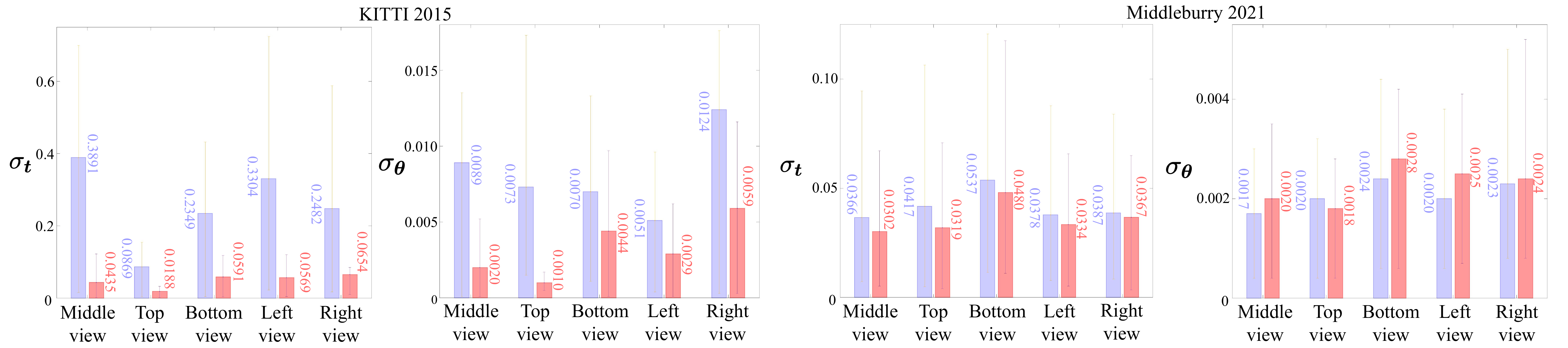}
\caption{
Comparison between \cite{ling2016high} and our proposed algorithm on the KITTI 2015 \cite{menze2015joint} and Middlebury 2021 \cite{scharstein2014high} datasets. {\protect\bariros} presents the results achieved by Ling and Shen \cite{ling2016high}, and {\protect\barours} presents the results achieved by our proposed method. 
}
\label{fig.KITTI_MiddleBurry}
\vspace{-0.5em}
\end{figure*}

\subsection{Comprehensive performance evaluation}
\label{sec.exp_results}

The quantitative experimental results on our created large-scale dataset are presented in Table \ref{tab.our_dataset_et_etheta} and Fig. \ref{fig.our_dataset_sigma}. These results suggest that our algorithm not only achieves higher accuracy in ${\boldsymbol{t}^*}$ and ${\boldsymbol{\theta}^*}$ estimation for multi-pair cases, but also demonstrates superior robustness when estimating ${\boldsymbol{t}^k}$ and ${\boldsymbol{\theta}^k}$ for single-pair cases, across various viewpoints within dynamic environments. Compared to the approach proposed in \cite{ling2016high}, our algorithm decreases $e_{\boldsymbol{\theta}}$ and $e_{\boldsymbol{t}}$ by an average of 47.63\% and 72.90\% in indoor scenarios. Additionally, it lowers $e_{\boldsymbol{\theta}}$ and $e_{\boldsymbol{t}}$ by an average of 72.78\% and 76.48\% in outdoor scenarios. In terms of $\sigma_{\boldsymbol{\theta}}$ and $\sigma_{\boldsymbol{t}}$, our algorithm lowers them by an average of 52.27\% and 77.36\% in indoor scenarios and by an average of 49.87\% and 55.17\% in outdoor scenarios when compared to the approach proposed in \cite{ling2016high}. 
However, the unsatisfactory performance of both algorithms in indoor environments has exceeded our initial expectations. Upon analysis, we attribute these unexpected results to the dynamic data collection process, in which we moved the stereo camera to capture video sequences for global optimum extrinsic parameter estimation. In indoor environments, the 3-D coordinates of detected keypoints are typically in close proximity to the stereo camera, leading to relatively large positional differences between successive video frames. Consequently, this can introduce notable motion blur in the images, resulting in less accurate matched correspondences.

\begin{figure*}[t]
\vspace{+0.5em}
\centering
\includegraphics[width=0.99\textwidth]{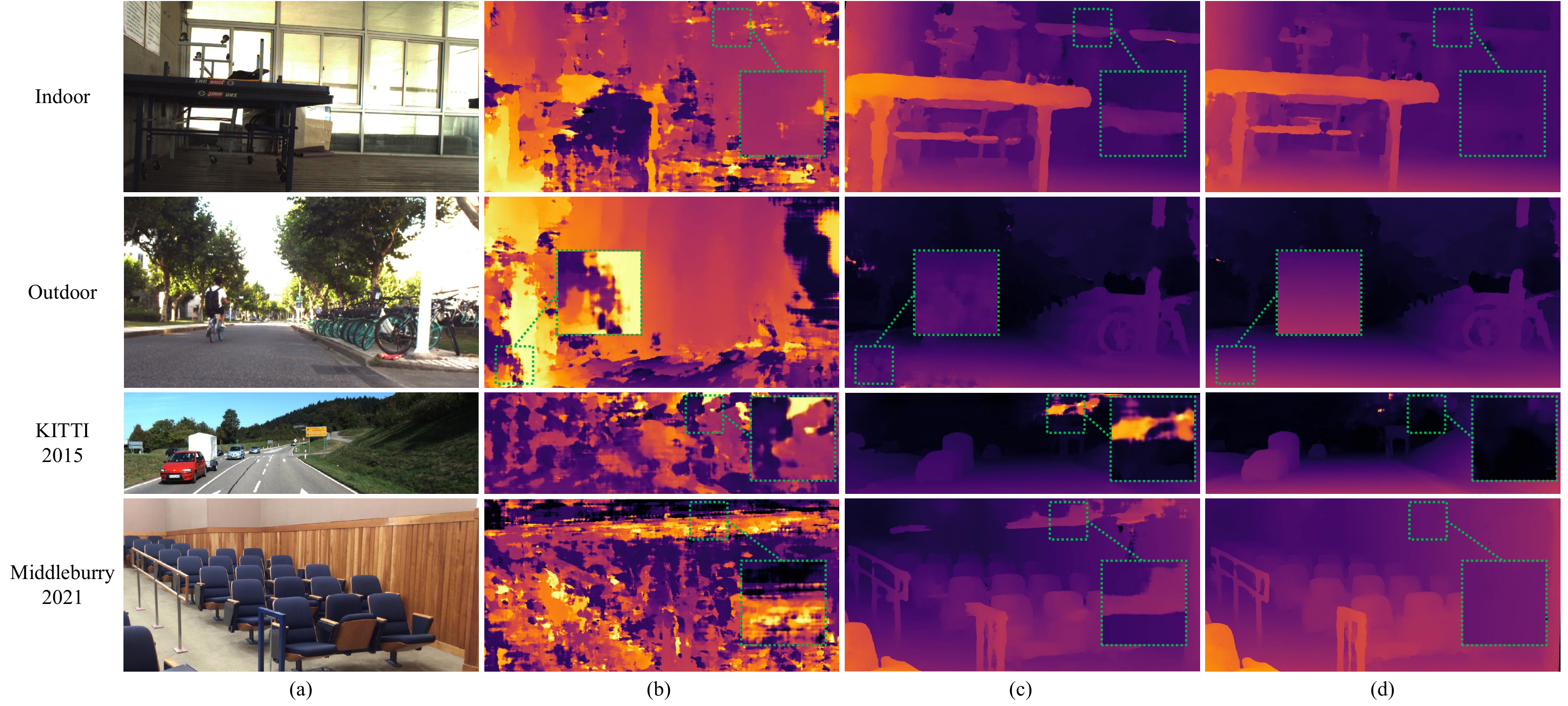}
\vspace{-0.5em}
\caption{
Qualitative experimental results of disparity estimation: (a) left images; (b) disparity maps estimated using unrectified stereo images; (c) disparity maps estimated using stereo images rectified based on the extrinsic parameters estimated using Ling and Shen's algorithm \cite{ling2016high}; (d) disparity maps estimated using stereo images rectified based on the extrinsic parameters estimated using our proposed algorithm. 
}
\label{fig.disp}
\end{figure*}

As can be seen from Table \ref{tab.kitti_middleburry_error} and Fig. \ref{fig.KITTI_MiddleBurry}, the quantitative experimental results on the KITTI 2015 dataset yield conclusions consistent with those observed in the above-mentioned outdoor experiments. We analyze that this might be due to the fact that image quality in the KITTI 2015 dataset is slightly higher than in our dataset, with fewer cases affected by motion blur, enabling both algorithms to achieve relatively stable results. Furthermore, due to the fact that moving vehicles generally have a negligible yaw angle, except for when they are turning, the estimation of the rotation vector is relatively stable and accurate. Through a comprehensive analysis of our algorithm's performance on both our created dataset and the KITTI 2015 dataset, we believe that our algorithm is less sensitive to image quality. It demonstrates the capability to provide feasible solutions even when images have motion-induced blur. On the other hand, the experimental results on the Middlebury 2021 dataset provide additional support for our perspective on algorithm performance in both static and dynamic environments. Our algorithm decreases $e_{\boldsymbol{\theta}}$ and $e_{\boldsymbol{t}}$ by an average of 35.62\% and 66.04\%, respectively. Our achieved $\sigma_{\boldsymbol{\theta}}$ and $\sigma_{\boldsymbol{t}}$ are comparable to those yielded by \cite{ling2016high}. 
This is attributed to the high quality and static nature of the images captured using a high-definition camera in the Middlebury 2021 dataset.

Moreover, we conduct additional qualitative experiments to provide a more comprehensive comparison of performance between the baseline algorithm \cite{ling2016high} and our proposed algorithm. As illustrated in Fig. \ref{fig.disp}, the quality of disparity maps estimated from unrectified stereo images is notably low, while disparity maps estimated using rectified stereo images have higher quality. Specifically, when employing our proposed algorithm to self-calibrate the stereo camera and rectify the stereo images, the resulting disparity maps demonstrate improved accuracy with fewer erroneous regions, as compared to those obtained using the baseline algorithm \cite{ling2016high}.

The most significant discovery in our experiments is the observed instability in translation vector estimation when using the algorithm presented in \cite{ling2016high}. Their approach raises concerns, as it produces a coefficient matrix that is nearly positive semidefinite around the optimum point. This characteristic can potentially lead to instability during the translation vector optimization process, particularly in cases where the detected keypoints are not sufficiently accurate, resulting in the failure of the iteration step size to converge to zero. In contrast, our algorithm focuses on the optimization of $\boldsymbol{R}_l$ and $\boldsymbol{R}_r$ rather than $\boldsymbol{R}$ and $\boldsymbol{t}$. This approach enhances the overall stability, even in scenarios where the quality of keypoints is unsatisfactory. This phenomenon is also evident in the Middlebury experiments, as illustrated in Fig. \ref{fig.KITTI_MiddleBurry}. Thanks to the high-definition images in the Middlebury dataset, which guarantee accurate correspondences between stereo images, both algorithms exhibit stable performance.

To further evaluate the accuracy of our algorithm and analyze the impact of correspondence matching on extrinsic calibration, we compute the reprojection error w.r.t. different numbers of correspondence pairs and the standard deviations of matching error. The experimental results are detailed in our supplement. We observe that when correspondence matching is reliable, the reprojection error is less than 1 pixel, demonstrating the high accuracy of our algorithm.

\section{Conclusion}
\label{sec.conclusion}
This paper presented two significant algorithmic contributions: (1) a stereo camera online self-calibration algorithm built upon the principles of stereo rectification for single-pair cases, and (2) an efficient and effective algorithm for globally optimizing extrinsic parameter estimation, when multiple stereo image pairs are available. In addition, this paper introduced four new, practical evaluation metrics to quantify the robustness and accuracy of extrinsic parameter estimation, applicable to both single-pair and multi-pair cases. Through comprehensive experiments conducted on our newly created indoor and outdoor datasets, as well as two public datasets, KITTI 2015 and Middlebury 2021, we demonstrated that the proposed algorithm significantly outperforms the state-of-the-art algorithm. With further optimization in algorithm efficiency, we are confident that the proposed algorithm can be incorporated into practical stereo vision systems to provide robust 3-D information for autonomous robots.

\clearpage
\normalem
\bibliographystyle{IEEEtran}
\bibliography{ref}
	
\end{document}